\pdfoutput=1

\documentclass[11pt]{article}

\usepackage[preprint]{acl}

\usepackage{times}
\usepackage{latexsym}
\usepackage{enumitem}
\usepackage{booktabs}
\usepackage{multirow}
\usepackage{color}

\usepackage[T1]{fontenc}

\usepackage[utf8]{inputenc}

\usepackage{microtype}
\usepackage{amsmath}

\usepackage{inconsolata}

\usepackage{graphicx}

\title{ChatMotion: A Multimodal Multi-Agent for Human Motion Analysis}

\begin{document}
\makeatletter
\def\thanks#1{\protected@xdef\@thanks{\@thanks
        \protect\footnotetext{#1}}}
\makeatother

\author{\textnormal{Lei Li\textsuperscript{\rm 1, 2, *, \dag}, 
Sen Jia\textsuperscript{\rm 3, *}, 
Jianhao Wang\textsuperscript{\rm 4}, 
Zhaochong An\textsuperscript{\rm 1}, }\\
Jiaang Li\textsuperscript{\rm 1}, 
Jenq-Neng Hwang\textsuperscript{\rm 2}, 
Serge Belongie\textsuperscript{\rm 1}
\thanks{\(^{*}\) These authors contributed equally to this work.} \thanks{\(^{\dag}\) Corresponding Author. (\href{mailto:lilei@di.ku.dk}{\color{black}{\texttt{lilei@di.ku.dk}}}) \textsuperscript{\rm 1} University of Copenhagen \textsuperscript{\rm 2} University of Washington \textsuperscript{\rm 3} Shandong University \textsuperscript{\rm 4} Xi'an Jiaotong University
}}

\maketitle
\begin{abstract}

Advancements in Multimodal Large Language Models (MLLMs) have improved human motion understanding. However, these models remain constrained by their "instruct-only" nature, lacking interactivity and adaptability for diverse analytical perspectives. To address these challenges, we introduce ChatMotion, a multimodal multi-agent framework for human motion analysis. ChatMotion dynamically interprets user intent, decomposes complex tasks into meta-tasks, and activates specialized function modules for motion comprehension. It integrates multiple specialized modules, such as the MotionCore, to analyze human motion from various perspectives. Extensive experiments demonstrate ChatMotion's precision, adaptability, and user engagement for human motion understanding.

\end{abstract}

\section{Introduction}



Human motion understanding~\cite{li2024human,zhou2024efficient,loper2023smpl,jiang2024back,jiang2023unihpe} has gained attention due to its wide-ranging applications in fields such as healthcare, human-computer interaction, rehabilitation, sports science, and virtual human modeling~\cite{plappert2016kit, zhang2021we, hong2022versatile, qu2024llms}. A deep understanding of human motion can drive advancements in areas like physical therapy~\cite{smeddinck2020human}, immersive virtual experiences~\cite{xiao2024study}, and assistive technology interfaces~\cite{khiabani2021semg}. As human motion data becomes more accessible, the demand for systems capable of effectively processing and analyzing this data has increased~\cite{zhang2024research}. However, existing motion understanding models often struggle to handle the accurate analysis of human motions and the dynamic nature of user requirements~\cite{meng2020recent, smeddinck2020human}. These MLLMs tend to exhibit limited adaptability to complex, multi-faceted user queries and are often constrained by biases inherent in single-model analyses~\cite{frangoudes2022assessing}, failing to integrate diverse insights into a comprehensive, generalizable, and accurate analysis~\cite{xu2021human}.

\begin{figure}[t]
\centering
\includegraphics[width=1\linewidth]{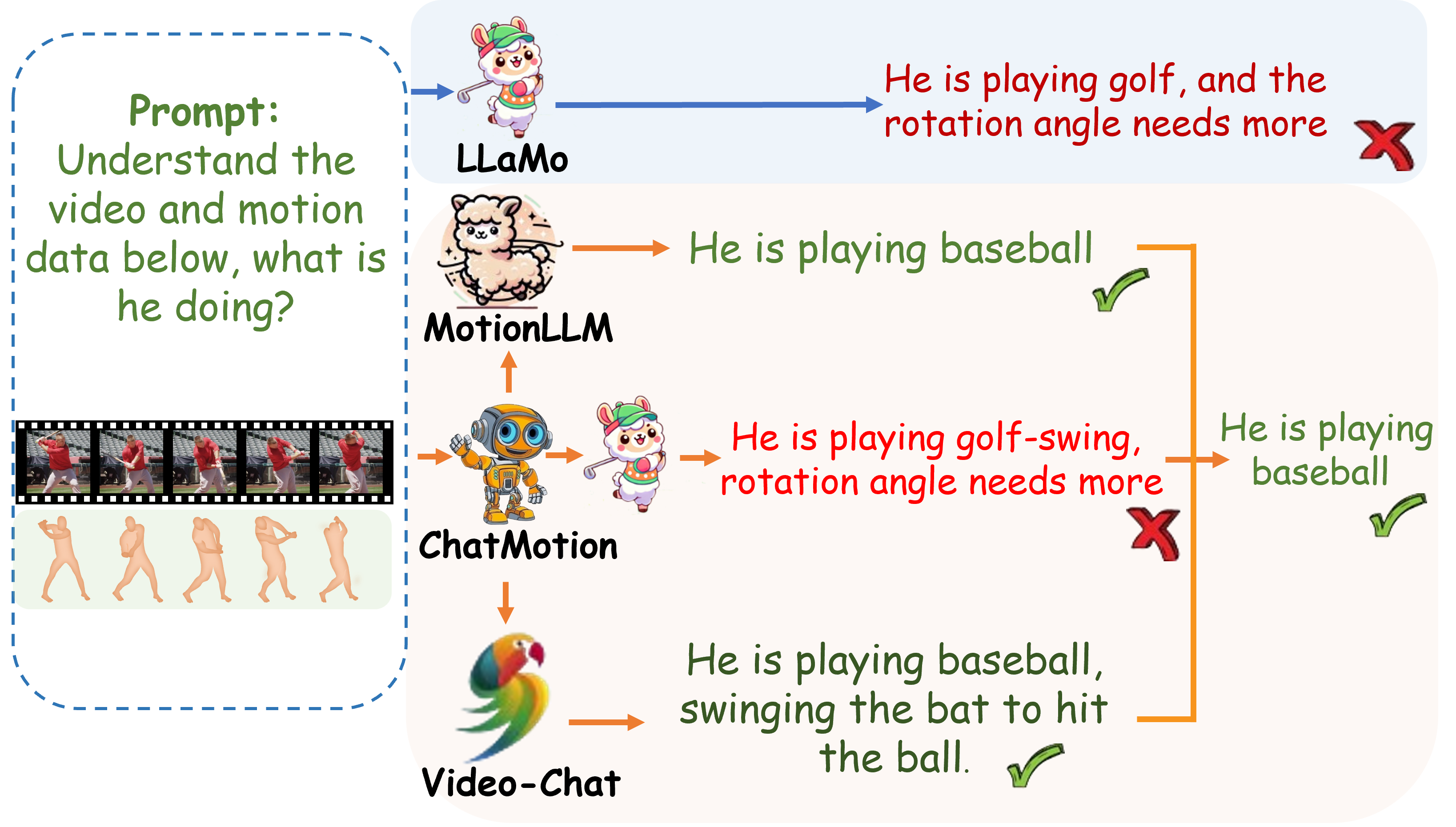} %
\vspace{-0.20in}
\caption{ChatMotion compares with LLaMo~\cite{li2024human}, a state-of-the-art MLLM for motion understanding. By integrating insights from multiple MLLM results, ChatMotion delivers more accurate analysis.}
\label{MotionCore}
\vspace{-0.22in}
\end{figure}

With the LLM-driven application development\cite{shi2025explaining,shi2024chops,cai2024t,liu2024graph,zheng2025reassessing,yang2024chain,cai2024role}, recent advancements in human motion understanding have progressed, particularly with LLM-based methods targeting specialized tasks and domain-specific applications. Models such as MotionGPT~\cite{jiang2023motiongpt} and MotionLLM~\cite{chen2024motionllm} propose methods to encode motion into structured formats, translating motion data (e.g., videos) into textual descriptions for general motion understanding tasks. Building on this foundation, LLaMo~\cite{li2024human} integrates a motion encoder and cross-talker without relying on motion quantification, demonstrating capabilities in general motion comprehension and specialized analysis across professional domains. These LLM-based motion models aim to bridge raw motion data and interpretable insights, enabling applications in diverse fields.

Despite these advancements, existing approaches still face limitations when applied to broader motion analysis tasks. A key challenge is their reliance on single-model architectures, which often struggle to address complex user requirements~\cite{wei2024motion}. These models show limited adaptability to dynamic user goals and lack mechanisms to integrate insights from multiple MLLMs, constraining their ability to provide comprehensive results. Additionally, they lack effective frameworks for verifying outcomes or refining analyses based on user feedback, which may affect reliability~\cite{lan2022analyzing}. As a result, current Motion LLMs encounter challenges in delivering accurate and complete human motion analyses.

To address these challenges and based on LLaMo \cite{li2024human}, we introduce \textbf{ChatMotion}, the first agent-based framework for motion understanding, combining multi-agent systems with the MotionCore toolbox. Given motion or video data with a user prompt, ChatMotion uses a planner to decompose the task into sub-tasks, which are then handled by the Executor using tools within MotionCore. The MotionCore consists of four modules: MotionAnalyzer, Aggregator, Generator, and Auxiliary Module. The Executor calls upon the MotionAnalyzer, utilizing multiple motion LLMs to analyze data from various perspectives. The Aggregator, with two mechanisms, synthesizes the most probable result from the MotionAnalyzer outputs. The Generator reviews the user's request and synthesizes the answer, leveraging contextual information from other modules. A verifier ensures consistency and relevance of intermediate results, enhancing the reliability of the final output. Through coordinated agent efforts, ChatMotion provides a flexible, precise, and reliable approach to motion analysis, overcoming the limitations of traditional motion LLMs.

We validate ChatMotion across a wide range of general human motion understanding datasets (e.g., Movid~\cite{chen2024motionllm}, BABEL-QA~\cite{endo2023motion}, MVbench~\cite{li2024mvbench}, and Mo-Repcount~\cite{li2024human} ), demonstrating its effectiveness across both standard and complex tasks. Experimental results highlight the improvements in accuracy, adaptability, and user engagement, establishing new benchmarks in the field of human motion analysis. In summary, the contributions of this work are as follows:
\begin{itemize}[noitemsep, topsep=0pt]
    \item \textbf{ChatMotion}, a multi-agent system with a planner-Executor-verifier architecture for comprehensive human motion analysis.
    \item A robust \textbf{MotionCore} for invoking functional tools to achieve advanced comprehension by synthesizing multiple perspectives from various MLLMs and can be readily extended, ensuring adaptability and scalability.
    \item Empirical validation across multiple datasets demonstrates that ChatMotion achieves improved performance in human motion analysis compared to existing MLLMs.
\end{itemize}

\begin{figure*}[ht]
\centering
\includegraphics[width=1\linewidth]{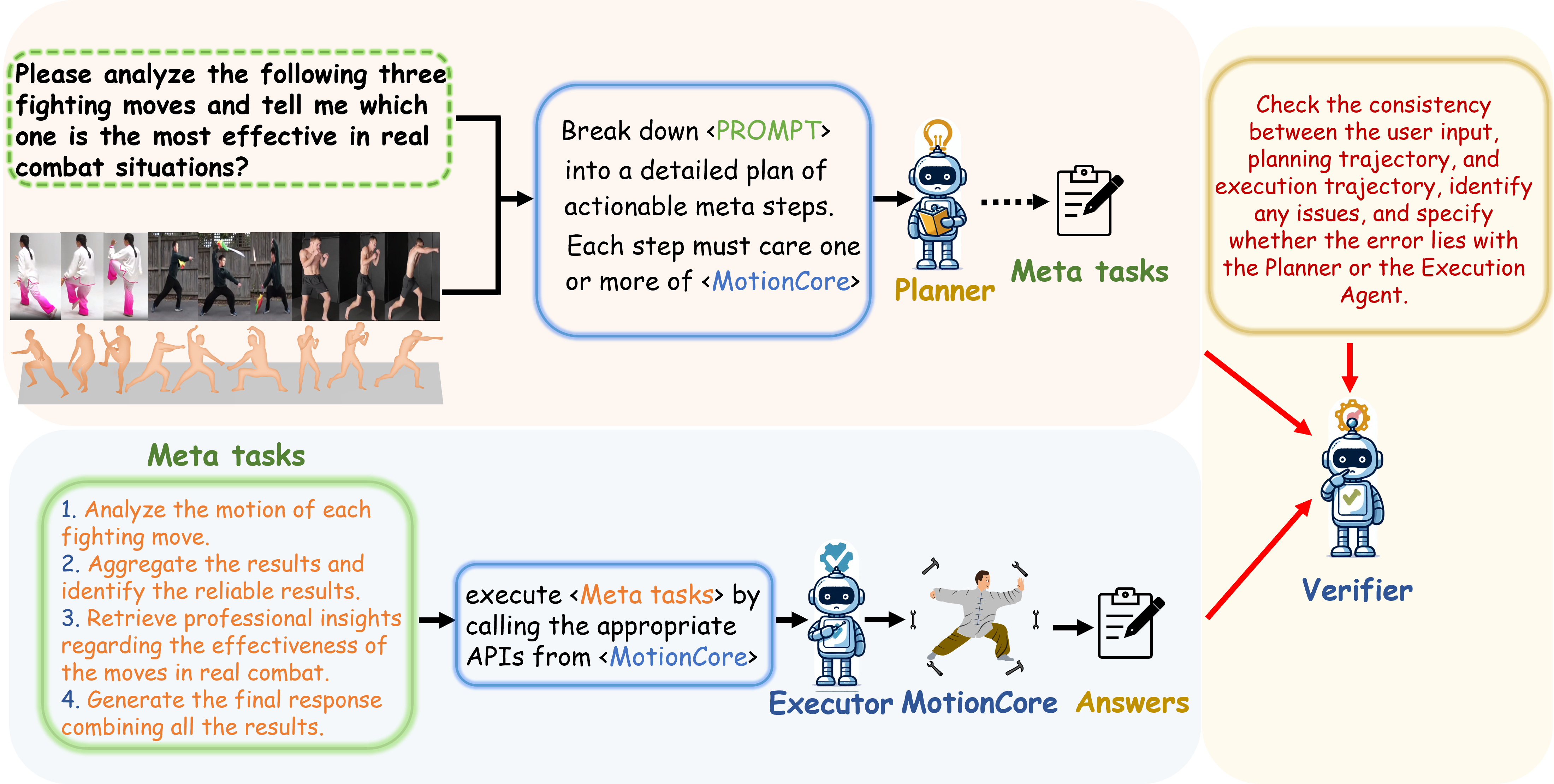} %
\vspace{-0.2in}
\caption{The ChatMotion pipeline operates through a three-stage framework designed to optimize task resolution. The Planner interprets the user’s query and breaks it into meta-tasks. Then, the Executor selects and applies appropriate MotionCore tools to execute these tasks. Finally, the Verifier ensures overall correctness, coherence, and completeness.}
\label{ChatMotion_pipeline}
\vspace{-0.15in}
\end{figure*}

\section{Related works}

\subsection{Human Multimodal Representations}

Multimodal representation learning is pivotal for human-centric analyses, especially in tasks requiring spatial-temporal reasoning to interpret complex behaviors~\cite{lin2023videollm, ning2023videobench, li2023videochat}. Recent advancements, such as Video-LLaVA, integrate visual information from images and videos into a unified linguistic feature space, enabling improved visual reasoning for behavioral analysis~\cite{lin2023videollm}. However, many models remain limited to isolated video frames and privacy concerns, constraining their effectiveness in the dynamic real world.~\cite{ning2023videobench, heilbron2015activitynet, maaz2023video}. To address these limitations, motion data has emerged as a privacy-preserving alternative, allowing action analysis without revealing identifiable visual details~\cite{song2023adaptive, yang2023recognizing}. By combining visual and motion data, emerging multimodal frameworks offer comprehensive, privacy-aware solutions, leveraging the complementary strengths of both modalities for enhanced adaptability across diverse applications.

\subsection{Human Motion Understanding}

Human motion analysis traditionally relies on skeletal data, represented as joint keypoint sequences, to capture movement dynamics while preserving user privacy~\cite{shi2023learning, plappert2018bidirectional, yang2023understanding}. Early methods, such as 2s-AGCN~\cite{shi2019two}, and recent transformer-based models like MotionCLIP~\cite{chen2023motiongpt}, have demonstrated success in tasks such as activity recognition, caption generation, and behavior analysis by translating motion data into language tokens. While effective in modeling structural movement patterns, these approaches often neglect environmental context, which is crucial for interpreting motions that may convey different meanings based on situational factors~\cite{song2023finegrained, maaz2023video}. To address this, recent models integrate motion and visual data, enabling improved generalization in dynamic and diverse environments~\cite{liu2024pose, he2023activitynet}. Frameworks like LLaMo\cite{li2024human} have further advanced the field by incorporating motion encoders, estimators, and efficient fusion mechanisms, achieving state-of-the-art results in both general and specialized motion analysis.

\vspace{-0.10in}
\begin{figure}[t]
\centering
\includegraphics[width=1\linewidth]{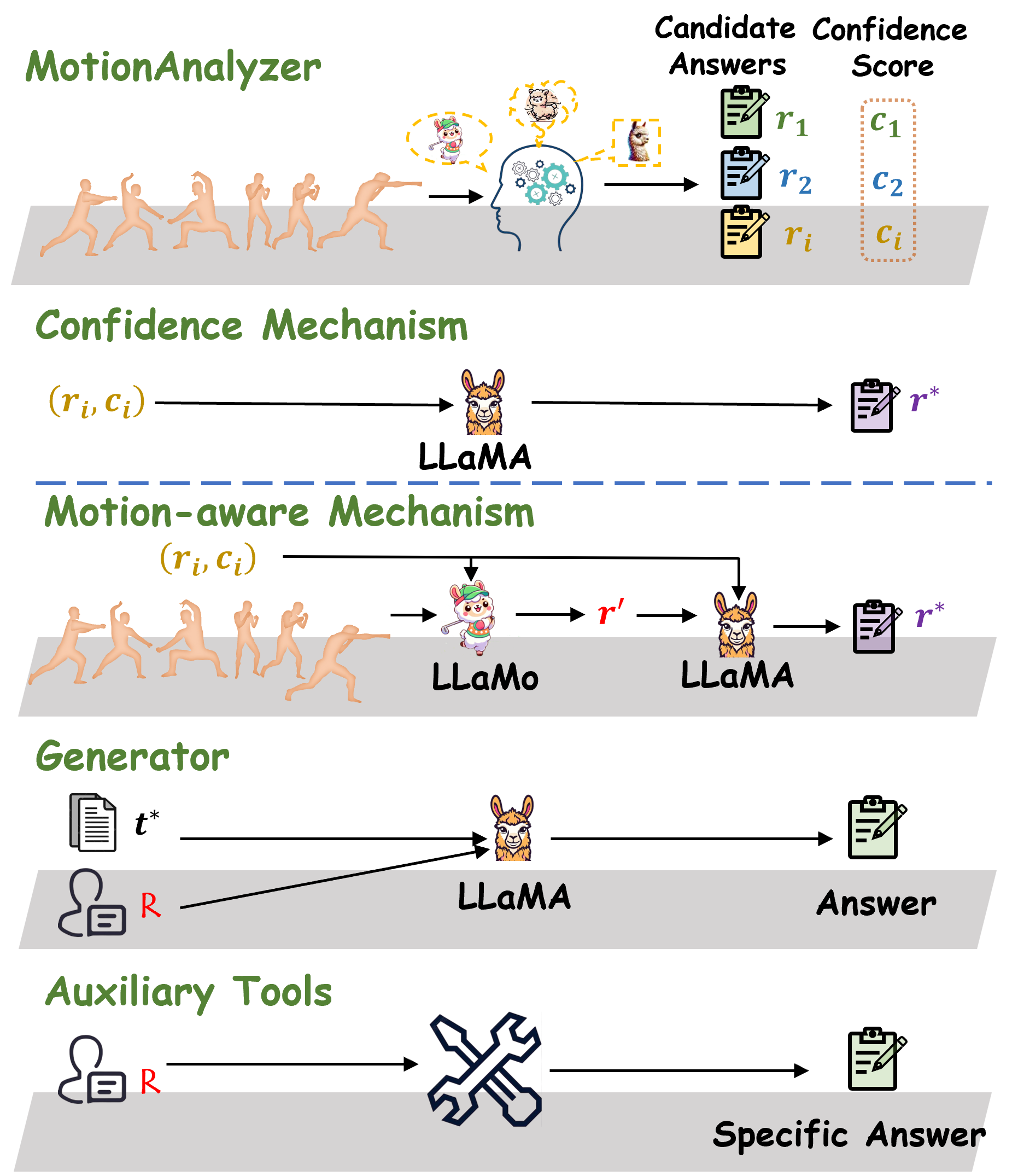} %
\vspace{-0.20in}
\caption{Components of MotionCore: the MotionCore integrates the MotionAnalyzer and Selection modules to concurrently process and aggregate multiple human motion analyses in two specific ways. The Generation Module synthesizes and contextualizes the results to align with user queries. Additionally, an auxiliary toolbox enables dynamic expansion with supplementary tools to address evolving user requirements.}
\label{MotionCore}
\vspace{-0.15in}
\end{figure}

\section{ChatMotion}


As shown in Fig.~\ref{ChatMotion_pipeline}, ChatMotion is a multi-agent system that processes user queries involving motion and video data through the Planner, Executor, and Verifier, with LLaMA-70B~\cite{touvron2023llama} employed for all agents. The Planner decomposes the task into meta-tasks, the Executor executes them via MotionCore function calls, and the Verifier ensures accuracy, delivering context-aware, precise results for complex motion analysis.


\subsection{Planner}
The planner serves as the decision-maker, interpreting user intent and subdividing complex tasks into structured meta-tasks. It first analyzes the input query to identify the core objectives and dependencies within the task, and then breaks the task down into smaller, manageable meta-tasks. It operates as the initial step in the multi-agent framework, ensuring that user requirements are translated into a structured workflow that aligns with evolving goals. 

Specifically, let us denote a user query by \(R\). As the simplified version is illustrated in Fig.~\ref{ChatMotion_pipeline}, the Planner will receive an instruction containing user query and available tools functionality in MotionCore which is a function toolbox tailored for human motion analysis (see Sec.~\ref{MotionCore}). Then, the Planner will follow the instructions and identify a set of core objectives \(\mathcal{O} = \{O_1, O_2, \ldots, O_m\}\) simply based on \(R\). 
These objectives are then decomposed into finer-grained meta-tasks guided by the specific functionalities available in the MotionCore tools. 
\[
\mathcal{M} = \{M_1, M_2, \ldots, M_k\},
\]
where each \(M_i\) represents a meta-task in the overall workflow. This decomposition allows the system to handle a wide range of user inputs, from simple queries to multi-step, dynamic tasks.

\subsection{Executor}
Executor serves as the core execution component, responsible for translating the Planner’s meta-tasks into actionable operations using a suite of function tools. After provided the meta-tasks \(\mathcal{M}\), the Executor will process each task in turn guided by the instruction as illustrated in Fig.~\ref{ChatMotion_pipeline}, determining and using the most appropriate function tools in MotionCore (see Sec.~\ref{MotionCore}) based on the alignment between their functional description and the objectives of the meta-task. 

Formally, for a given meta-task \(M_i \in \mathcal{M}\), The Executor will traverse functions capabilities within MotionCore and choose an appropriate tool \(\phi_i\) from a function tool set \(\Phi = \{\phi_1, \phi_2, \ldots, \phi_s\}\) in MotionCore, according to a mapping
\[
\Phi(M_i) \; \rightarrow \; \phi_i,
\]
where \(\phi_i\) is the specific function tool that best addresses the requirements of meta-task \(M_i\). 

If any meta-task proves infeasible, e.g., due to missing functionality, the Executor returns complete error information to the Planner, which will then update its tasks accordingly. The Executor reattempts these updated tasks, iterating through multiple rounds until the overall complex objective is met.


\subsection{Verifier}
\label{verifier}

The Verifier acts as a supervisory agent, ensuring the accuracy and reliability of the multi-agent workflow. It has two main roles: first, it checks that the Planner’s meta-tasks are logically structured and aligned with the user’s prompt; second, it verifies that the meta-tasks can be executed using available tools and that the results meet expectations. If any meta-task cannot be executed or produces incorrect results, or if the Executor calls an inappropriate function, the Verifier prompts the Planner to revise the task list or the Executor to select a different tool. This feedback loop ensures that tasks are executed correctly using the right tools.

\subsection{MotionCore}
\label{MotionCore}

MotionCore is a comprehensive toolkit that enables efficient human motion understanding by integrating various modules and auxiliary functions. It also includes auxiliary tools for tasks like motion visualization and video retrieval, meeting users' diverse requirements. MotionCore is orchestrated by the Executor Agent, which autonomously selects the appropriate tools from the toolkit to complete tasks based on a given meta-task list.

\subsubsection{MotionAnalyzer}

The  MotionAnalyzer in MotionCore enhances motion understanding and mitigates biases through a dynamic, multi-model approach. It integrates human motion models, such as MotionLLM~\cite{chen2024motionllm}, MotionGPT~\cite{jiang2023motiongpt}, and LLaMo\cite{li2024human}, alongside video captioning models such as VideoChat2~\cite{li2024mvbench}, GPT-4v~\cite{openai20234v}, and video-LLaVA~\cite{lin2023video} to handle human motion input.

Let the set of motion understanding models be denoted as $\{F_1, F_2, \ldots, F_N\}$, where each model $F_i$ processes the multimodal input data $D$ (e.g., video frames, motion capture data) to produce text analysis $r_i$, i.e., $(r_i) = F_i(D), \quad i = 1, 2, \ldots, N$. Each model is assigned a predefined confidence score $c_i$, based on the previous evaluation performance, independent of the model’s predictions. These confidence scores are allocated based on the input modalities, which can be motion capture, video, or motion-video. The outputs and their corresponding confidence scores are represented as $\{(r_1, c_1), (r_2, c_2), \ldots, (r_N, c_N)\}$, where $c_i$ denotes the predefined confidence score for the output $r_i$ of model $F_i$ in its respective task. This integration of predefined confidence scores ensures a robust and flexible understanding of motion, leveraging the strengths of each model across diverse modalities and tasks.

\subsubsection{Aggregator}  
The Aggregator in MotionCore identifies the most reliable result from a set of \(\{(r_i, c_i)\}\) pairs, employing two strategies: the Confidence Mechanism and the Motion-aware Mechanism, which enhance the robustness of motion understanding by selecting the most accurate outcome from diverse perspectives.

\paragraph{Confidence Mechanism}  
Rooted in game theory, this method considers the set  
\[
\{ (r_i, c_i) \mid i = 1, 2, \dots, N \},
\]
where \(r_i\) is a model’s output and \(c_i\) is its associated confidence score. The mechanism assigns higher weight to more confident outputs, with a "majority wins" principle when models converge on similar results. Rather than using a fixed function, the analysis-confidence pairs \(\{(r_i, c_i)\}\) are passed to LLaMA~\cite{touvron2023llama}, which adaptively integrates the outputs by balancing consensus with individual model expertise. This ensures a flexible and robust aggregation process, emphasizing shared conclusions while considering outlier predictions.

Though foundational, this approach is basic, relying primarily on confidence scores and model consensus. The next step incorporates a motion-aware mechanism to refine the process.

\paragraph{Motion-aware Mechanism}  
With LLaMo's~\cite{li2024human} specialized motion-understanding capabilities, this mechanism evaluates \(\{(r_i, c_i)\}\) pairs alongside the original motion or video data \(\mathcal{M}\), generating an initial estimate:
\[
r' = \mathrm{LLaMo}(r_1, \ldots, r_N;\; c_1, \ldots, c_N;\; \mathcal{M}).
\]
LLaMA~\cite{touvron2023llama} then re-examines the preliminary result \(r'\) and the original pairs \(\{(r_i, c_i)\}\) to mitigate model bias and refine the outcome. This dual-layer evaluation leverages LLaMo's domain-specific motion expertise and LLaMA's context-aware reasoning, improving both reliability and precision.

The Aggregator is a powerful tool within MotionCore, enabling ChatMotion to identify the most accurate analyses from diverse model outputs, fostering a more comprehensive understanding of human motion.


\subsubsection{Generator}
In MotionCore, the Generator is responsible for synthesizing contextual information from previous function calls and the user's original request to produce a final answer. As illustrated in Fig.~\ref{MotionCore}, the Generator reviews the user query and organizes the context into a coherent and accurate answer. The answer could be in the form of textual analysis, motion feedback, or other formats, depending on the user's request. Contextual information from earlier interactions is denoted as \(t^*\). The module then integrates this context with the user’s specific requirements, represented as \(R\), to generate a comprehensive response:
\[
\mathrm{Answer} = \Gamma(t^*, R),
\]
where \(\Gamma(\cdot)\) denotes LLaMA~\cite{touvron2023llama} by default. The purpose of the Generator is to transform the context into an answer that directly addresses the user’s needs, ensuring the answer is concise and contextually accurate.

\subsubsection{Auxiliary Tools} 
The Auxiliary Tools in MotionCore, which can be accessed by the Executor, extend ChatMotion's capabilities by orchestrating external, domain-specific functionalities that go beyond the scope of the multimodal model alone. For instance, the system can retrieve professional analysis by querying specialized knowledge bases, which provide context-specific insights based on user inputs. Additionally, it enables motion retrieval by identifying relevant motion data based on the user’s request, leveraging a stored database of labeled motion data and utilizing vector-based search to match the query to the most relevant motion. As a result, it equips ChatMotion with diverse motion analysis capabilities that simple MLLMs do not possess. By offering a unified, modular interface for diverse auxiliary function calls, ChatMotion readily integrates and extends new capabilities without overburdening the core model.

\begin{table*}[ht]
\centering
\renewcommand{\arraystretch}{0.8} 

\small
\resizebox{\textwidth}{!}{
\begin{tabular}{l|cc|cc|cc|cc|cc|cc|cc}
\toprule
\textbf{MoVid-Bench-Motion} & \multicolumn{2}{c|}{\textbf{Body.}} & \multicolumn{2}{c|}{\textbf{Seq.}} & \multicolumn{2}{c|}{\textbf{Dir.}} & \multicolumn{2}{c|}{\textbf{Rea.}} & \multicolumn{2}{c|}{\textbf{Hall.}} & \multicolumn{2}{c}{\textbf{All}} \\
& \textbf{Acc.} & \textbf{Score} & \textbf{Acc.} & \textbf{Score} & \textbf{Acc.} & \textbf{Score} & \textbf{Acc.} & \textbf{Score} & \textbf{Acc.} & \textbf{Score} & \textbf{Acc.} & \textbf{Score} \\ 
\midrule
GT & \textbf{100.00} & \textbf{5.00} & \textbf{100.00} & \textbf{5.00} & \textbf{100.00} & \textbf{5.00} & \textbf{100.00} & \textbf{5.00} & \textbf{100.00} & \textbf{5.00} & \textbf{100.00} & \textbf{5.00} \\
GPT-3.5~\cite{openai2023gpt35} & 24.51 & 2.04 & 30.41 & 2.25 & 27.14 & 2.19 & 39.19 & 2.64 & 58.33 & 3.22 & 31.33 & 2.31 \\
MotionGPT~\cite{jiang2023motiongpt} & 31.22 & 3.98 & 42.69 & \textbf{3.16} & 44.29 & 3.50 & 35.81 & 3.06 & 16.66 & 2.25 & 36.86 & 3.11 \\
MotionLLM~\cite{chen2024motionllm} & 50.49 & 3.55 & 36.84 & 3.14 & 58.57 & 3.76 & 52.70 & 3.58 & 55.56 & 3.39 & 49.50 & 3.49 \\
LLaMo~\cite{li2024human} & 59.30 & 4.01 & 44.01 & 3.12 & 60.91 & 3.99 & 58.21 & 3.64 & 61.17 & 3.53 & 55.32 & 3.67 \\
\midrule
\textbf{ChatMotion(CB)} & \textbf{60.89} & 4.03 & 46.21 & \textbf{3.30} & 62.11 & 4.03 & 59.53 & 3.77 & 68.95 & 3.78 & 56.90 & 3.72 \\
\textbf{ChatMotion} & 60.43 & \textbf{4.08} & \textbf{46.56} & 3.28 & \textbf{64.21} & \textbf{4.11} & \textbf{60.58} & \textbf{3.87} & \textbf{70.39} & \textbf{3.82} & \textbf{58.79} & \textbf{3.80} \\
\midrule
\textbf{MoVid-Bench-Video} & \multicolumn{2}{c|}{\textbf{Body.}} & \multicolumn{2}{c|}{\textbf{Seq.}} & \multicolumn{2}{c|}{\textbf{Dir.}} & \multicolumn{2}{c|}{\textbf{Rea.}} & \multicolumn{2}{c|}{\textbf{Hull.}} & \multicolumn{2}{c}{\textbf{All}} \\
& \textbf{Acc.} & \textbf{Score} & \textbf{Acc.} & \textbf{Score} & \textbf{Acc.} & \textbf{Score} & \textbf{Acc.} & \textbf{Score} & \textbf{Acc.} & \textbf{Score} & \textbf{Acc.} & \textbf{Score} \\ 
\midrule
GT & \textbf{100.00} & \textbf{5.00} & \textbf{100.00} & \textbf{5.00} & \textbf{100.00} & \textbf{5.00} & \textbf{100.00} & \textbf{5.00} & \textbf{100.00} & \textbf{5.00} & \textbf{100.00} & \textbf{5.00} \\
GPT-3.5~\cite{openai2023gpt35} & 2.40 & 1.23 & 1.39 & 1.00 & 4.65 & 1.09 & 5.41 & 1.65 & 0.00 & 0.94 & 3.03 & 1.26 \\
Video-LLAVA~\cite{lin2023video} & 33.53 & 2.76 & 25.46 & 2.72 & 41.86 & 2.84 & 52.97 & 3.28 & 58.83 & 1.89 & 42.53 & 2.70 \\
MotionLLM~\cite{chen2024motionllm} & 34.13 & 2.93 & 32.87 & 2.92 & 44.18 & 3.14 & 63.20 & 3.55 & 70.59 & 2.30 & 49.00 & 2.97 \\
LLaMo~\cite{li2024human} & 33.83 & 2.85 & 36.01 & 3.11 & 45.50 & 3.32 & 67.59 & 3.73 & 72.81 & 2.25 & 52.33 & 3.10 \\
\midrule
\textbf{ChatMotion(CB)} & \textbf{38.31} & \textbf{3.40} & 36.80 & 3.17 & 47.22 & 3.59 & 70.89 & 3.85 & 73.22 & \textbf{2.35} & 53.51 & 3.19 \\
\textbf{ChatMotion} & 38.06 & 3.34 & \textbf{37.39} & \textbf{3.18} & \textbf{47.92} & \textbf{3.65} & \textbf{72.16} & \textbf{3.99} & \textbf{74.01} & 2.30 & \textbf{54.96} & \textbf{3.25} \\
\bottomrule
\end{tabular}
}
\caption{Comparison between ChatMotion and existing Motion LLMs on the MoVid-Bench. The top part of the table presents motion-related results, and the bottom part presents video-related results. Higher accuracy and score values indicate better performance.}
\label{table:evaluation_on_movid}
\end{table*}

\begin{table*}[t]
\centering
\renewcommand{\arraystretch}{0.8} 
\setlength{\tabcolsep}{6pt}      
\scriptsize                      
\resizebox{\linewidth}{!}{      
\begin{tabular}{l|c|cccc|ccc}
\toprule
\textbf{Model} & \textbf{Pred. type} & \textbf{Overall} $\uparrow$ & \textbf{Action} $\uparrow$ & \textbf{Direction} $\uparrow$ & \textbf{Body Part} $\uparrow$ & \textbf{Before} $\uparrow$ & \textbf{After} $\uparrow$ & \textbf{Other} $\uparrow$ \\ 
\midrule
MotionCLIP-M~\cite{tevet2022motionclip} & cls. & 0.430 & 0.485 & 0.361 & 0.272 & 0.372 & 0.321 & 0.404\\ 
MotionCLIP-R~\cite{tevet2022motionclip} & cls. & 0.420 & 0.489 & 0.310 & 0.250 & 0.398 & 0.314 & 0.387\\ 
MotionLLM~\cite{chen2024motionllm} & gen. & 0.436 & 0.517 & 0.354 & 0.154 & 0.427 & 0.368 & 0.529 \\ 
LLaMo~\cite{li2024human} & gen. & 0.458 & 0.525 & 0.398 & 0.224 & 0.443 & 0.392 & 0.518 \\ 
\midrule
\textbf{ChatMotion(CB)} & gen. & 0.467 & 0.534 & 0.410 & \textbf{0.272} & 0.445 & 0.396 & 0.536 \\
\textbf{ChatMotion} & gen. & \textbf{0.473} & \textbf{0.537} & \textbf{0.412} & 0.265 & \textbf{0.451} & \textbf{0.406} & \textbf{0.537} \\
\bottomrule
\end{tabular}
} 
\caption{Comparison on BABEL-QA dataset. Higher scores indicate better performance. The results for ChatMotion's two methods are also included.}
\label{table:babel_qa_comparison}
\end{table*}

\section{Experimental Setup}

\paragraph{Datasets}
We evaluate ChatMotion on general human motion understanding benchmarks including Movid-bench~\cite{chen2024motionllm}, BABEL-QA~\cite{endo2023motion} and MVbench~\cite{li2024mvbench}, as well as Mo-Repcount~\cite{li2024human} for fine-grained motion capture capabilities. MoVid-Bench specifically assesses the model's ability to understand human behavior in both motion and video contexts. It consists of 1,350 data pairs, with 700 motion and 650 video samples, covering diverse daily scenarios in real-world. In addition, ChatMotion is tested on BABEL-QA and MVbench to evaluate motion-based and video-based question answering respectively. 

\paragraph{Tasks and Metrics}
ChatMotion is evaluated on tasks including action recognition, motion reasoning, and question answering. For MoVid-Bench, we follow established LLM evaluation metrics, assessing body-part recognition, sequential analysis, directionality, reasoning, and hallucination control in both motion and video contexts. BABEL-QA uses similar metrics with a focus on motion-related question answering, while Mo-Repcount employs specialized metrics like OBO, MAE, OBZ, and RMSE for fine-grained motion tracking accuracy. In the MVBench video understanding evaluation, we respond to multiple-choice questions by selecting the most suitable option as outlined in.

\paragraph{Baselines}
For our baselines, we select SoTA Motion LLMs for human-centric motion understanding, e.g., LLaMo~\cite{li2024human}, MotionLLM~\cite{chen2024motionllm} and MotionGPT~\cite{jiang2023motiongpt}. These models are widely recognized for their ability to process and understand human motion in both video and action contexts. For ChatMotion, \textbf{ChatMotion(CB)} and \textbf{ChatMotion} denote the versions using confidence-based and motion-aware aggregation, respectively. Through extensive comparison, our results highlight ChatMotion's exceptional ability to handle complex human motion understanding tasks, outperforming the selected baselines across a range of evaluation metrics.

\begin{figure*}[h]
    \centering
    \includegraphics[width=1\textwidth]{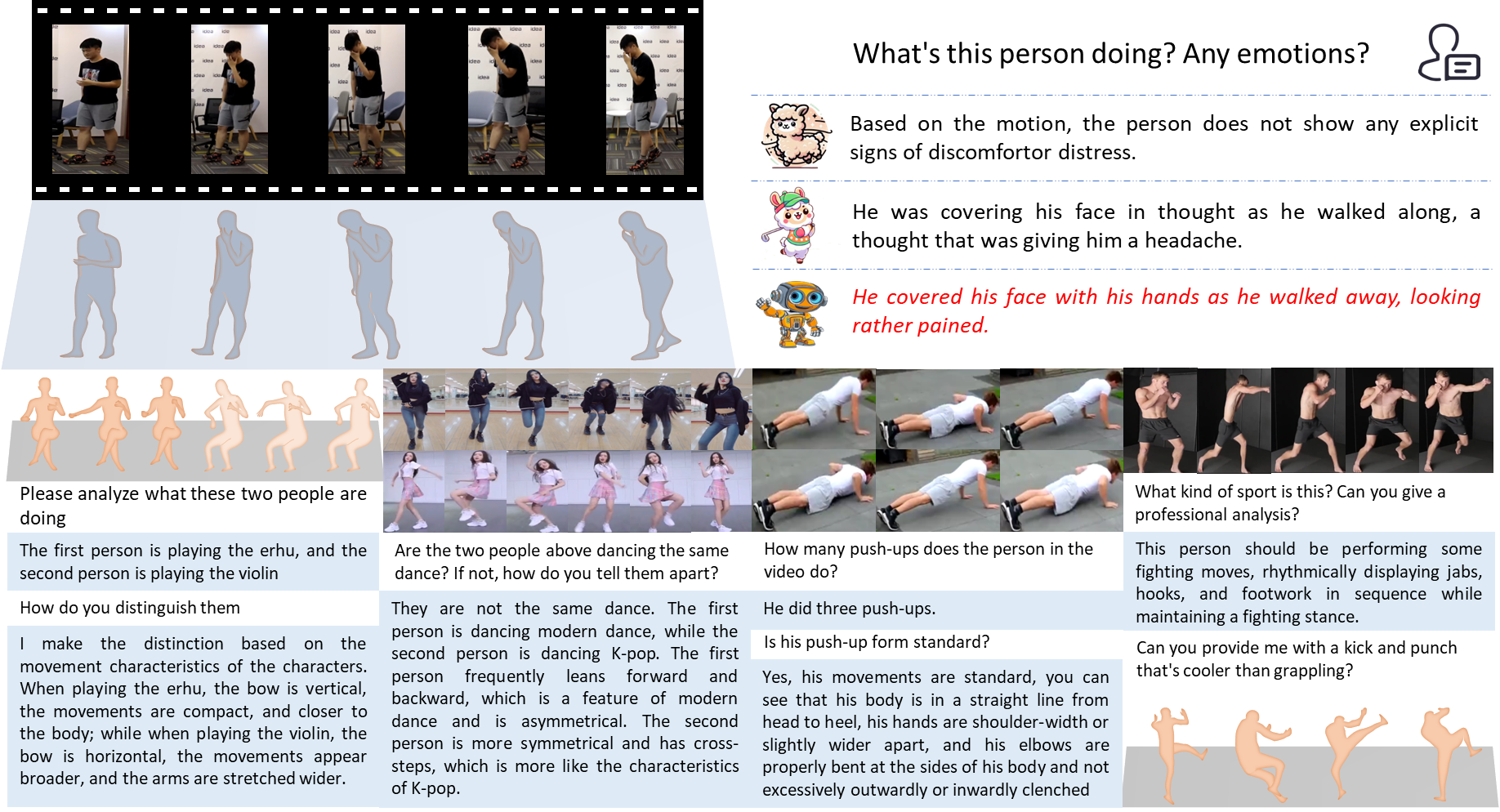}
    \caption{Examples of ChatMotion's responses in various human activities and sports, demonstrating its reasoning skills and specialized knowledge in active, movement-heavy scenarios.}
    \label{fig:qualitative}
\end{figure*}

\begin{table*}[ht]
\centering
\renewcommand{\arraystretch}{0.8} 
\setlength{\tabcolsep}{6pt}      
\scriptsize 
\resizebox{\linewidth}{!}{
\begin{tabular}{l|l|c|c|c|c|c|c|c|c|c}
\hline
\textbf{Model} & \textbf{LLM} & \textbf{Frames} & \textbf{AL} & \textbf{AP} & \textbf{AS} & \textbf{EN} & \textbf{FA} & \textbf{FP} & \textbf{UA} & \textbf{Avg.} \\
\hline
\text{Otter-V} & \text{Llama-7B} & 16 & 23.5 & 23.0 & 23.0 & 23.5 & 27.0 & 22.0 & 29.5 & 24.5 \\
\text{mPLUG-Owl-V } & \text{Llama-7B} & 16 & 23.0 & 28.0 & 22.0 & 26.0 & 29.0 & 24.0 & 29.0 & 25.8 \\
\text{VideoChatGPT } & \text{Vicuna-7B} & 100 & 20.0 & 26.0 & 23.5 & 29.5 & 22.5 & 22.5 & 29.0 & 25.2 \\
\text{VideoLLaMA } & \text{Vicuna-7B} & 16 & 22.5 & 25.5 & 27.5 & 30.0 & 29.0 & 32.5 & 39.0 & 29.4 \\
\text{VideoChat} & \text{Vicuna-7B} & 16 & 27.0 & 26.5 & 33.5 & 23.5 & 33.5 & 26.5 & 40.5 & 30.1 \\
\text{Video-LLAVA} & \text{Vicuna-7B} & 8 & 22.5 & 25.5 & 29.5 & 29.0 & 24.5 & 28.5 & 24.5 & 26.3 \\
\text{GPT-4v} & \text{GPT-4} & 16 & 40.5 & 63.5 & 55.5 & 31.0 & 46.5 & 47.5 & 73.5 & 51.1 \\
\text{VideoChat2} & \text{Vicuna-7B} & 16 & 23.0 & \textbf{66.0} & 47.5 & \textbf{35.0} & \textbf{49.5} & 49.0 & 60.0 & 47.1 \\
\text{MotionLLM} & \text{Vicuna-7B} & 8 & 33.0 & 29.5 & 32.5 & 29.0 & 31.5 & 28.5 & 37.5 & 31.6 \\
\hline
\textbf{ChatMotion(CB)} & \textbf{Agent} & \textbackslash & 42.0 & 65.5 & 56.0 & 33.0 & 48.0 & 50.5 & 72.0 & 52.4 \\
\textbf{ChatMotion} & \textbf{Agent} & \textbackslash & \textbf{43.0} & 65.5 & \textbf{58.0} & 34.0 & 49.0 & \textbf{51.0} & \textbf{74.0} & \textbf{53.2} \\
\hline
\end{tabular}
}
\caption{Performance of various models across different metrics, including GPT-4v, VideoChat2, MotionLLM and ChatMotion.}
\label{MVBench}
\end{table*}

\begin{table}[h!]
\small 
\centering
\begin{tabular}{p{2.2cm}|p{0.8cm} p{0.8cm} p{0.8cm} p{0.8cm}}
\toprule
\textbf{Model}   & \textbf{OBO} & \textbf{MAE} & \textbf{OBZ} & \textbf{RMSE} \\ \midrule
EScounts         & 0.397        & 0.291        & 0.198        & 5.58          \\
PoseRAC         & 0.382        & 0.312        & 0.204        & 5.95          \\
TransRAC         & 0.276        & 0.444        & 0.105        & 8.56          \\
RepNet          & 0.009        & \textbackslash          & \textbackslash            & \textbackslash             \\ 
MotionLLM      & 0.011   & \textbackslash   & \textbackslash   & \textbackslash  \\
LLaMo   & 0.389 & 0.324 & 0.222 & 6.15 \\
\midrule
\textbf{ChatMotion(CB)} & \textbf{0.412} & 0.279 & 0.229 & 5.33 \\
\textbf{ChatMotion} & 0.410 & \textbf{0.271} & \textbf{0.240} & \textbf{5.21} \\
\bottomrule
\end{tabular}
\caption{Motion and video details capture evaluation on Mo-RepCount.}

\label{table:mo-repcount-evaluation}
\vspace{-0.22in}
\end{table}

\section{Results}

\subsection{Quantitative Analysis}

\paragraph{Evaluation on Motion Understanding in MoVid-Bench.}

Table~\ref{table:evaluation_on_movid} compares the performance of motion-based LLMs on MoVid-Bench-Motion. Both ChatMotion(CB) and ChatMotion outperform existing baselines across all metrics. ChatMotion achieves an accuracy of 58.79\% and a score of 3.80, surpassing LLaMo by 3.47\% in accuracy and 0.13 in score. It also demonstrates strong hallucination control, achieving 70.39\% accuracy compared to LLaMo’s 61.17\%, underscoring the effectiveness of ChatMotion’s multi-model integration via its robust selection strategy. 

Previous models, such as MotionLLM and MotionGPT, lose fine-grained motion details due to motion discretization, leading to lower performance. Although LLaMo improves motion encoding, its single LLM-based structure introduces biases that limit its motion understanding capabilities. In contrast, ChatMotion leverages multi-agent collaboration and multi-model aggregation to enhance motion understanding. This approach reduces biases inherent in single LLM-based motion models and improves performance in motion sequence analysis. By integrating multiple agents, ChatMotion achieves greater robustness, demonstrating its superior capabilities to capture diverse motion dynamics and delivers more accurate, reliable results in complex motion understanding tasks.

\paragraph{Evaluation on Video Understanding in MoVid-Bench.}
ChatMotion(CB) demonstrates improvements across multiple metrics on MoVid-Bench-Video as shown in Table~\ref{table:evaluation_on_movid}, achieving an overall accuracy of 53.51\% and a score of 3.19, surpassing baseline models in all evaluated tasks. This performance gain is due to its effective aggregation of diverse video analysis perspectives, combined with confidence scores to ensure more reliable and stable reasoning. Furthermore, ChatMotion, with its motion-aware mechanism, further refines the analysis by better handling motion-related tasks, surpassing ChatMotion(CB) with an accuracy improvement of 1.45\% and a score increase of 0.06. This enhancement allows it to more effectively aggregate and analyze motion data, pushing performance beyond that of standard models. These innovations in model design, coupled with the synergistic effects of specialized modules, allow ChatMotion(CB) and ChatMotion to set new benchmarks in multimodal human motion analysis, outperforming existing LLM-based motion models across multiple tasks and metrics.

\paragraph{Evaluation on BABEL-QA.}
We evaluated ChatMotion on the BABEL-QA dataset to assess its performance in responding to complex motion-based queries. As shown in Table \ref{table:babel_qa_comparison}, both ChatMotion(CB) and ChatMotion outperform other LLM-based motion models across several metrics. ChatMotion(CB) achieves an overall score of 0.467, while ChatMotion further improves this to 0.473, demonstrating its enhanced capability. This improvement is due to ChatMotion's motion-aware mechanism, which takes both motion inputs and candidate results into account. By leveraging LLaMo's advanced multimodal capabilities, ChatMotion esures more robust and stable results. Despite some limitations on specific metrics, ChatMotion compensates for these and delivers superior overall results. These advancements position ChatMotion as a new benchmark in motion-based question answering, highlighting the effectiveness of multimodal aggregation and motion-aware mechanisms in achieving more accurate and reliable results.

\paragraph{Evaluation on MVBench.}
We evaluated ChatMotion on the MVBench dataset to assess its performance in video question answering across seven motion understanding sub-tasks. As shown in Table~\ref{MVBench}, ChatMotion(CB) outperforms MotionLLM~\cite{chen2024motionllm}, the LLM-based motion understanding model, achieves an average score of 52.4, while ChatMotion increases this to 53.2. These results highlight the efficacy of ChatMotion’s multi-agent framework, which reduces biases inherent to LLM-based motion models by incorporating dynamic function calls. Performance gains are particularly evident in most metrics, demonstrating the advantages of multi-agent integration for robust motion understanding. While slight performance gaps persist in specific tasks compared with expert models (e.g., EN of VideChat2), the overall improvement over the LLM-based motion model, MotionLLM, remains statistically better.

\paragraph{Evaluation on Mo-Repcount}
To evaluate ChatMotion's performance on fine-grained motion tasks, we benchmarked it on Mo-Repcount against SoTA Motion LLMs. The results in Table~\ref{table:mo-repcount-evaluation} show that ChatMotion outperforms LLaMo by 4\%-8\% across all metrics, demonstrating ChatMotion's advanced capability to aggregate the strengths of specialized models and achieve superior performance in fine-grained motion tasks.



\subsection{Qualitative Analysis}

Qualitative results, as shown in Fig.~\ref{fig:qualitative}, demonstrate ChatMotion's superior capabilities in understanding human motion across diverse scenarios. In a task where a human expresses sadness, using both video and motion inputs, MotionLLM fails to provide a correct interpretation, while LLaMo identifies the emotion, though with some ambiguity. Notably, ChatMotion excels in tasks that current LLM-based motion models struggle with, including fine-grained counting and comprehensive analyses utilizing RAG, alongside detailed comparisons of motion-capture and video data. These results showcase the model's ability to handle complex, multimodal motion tasks that require context-sensitive reasoning beyond the capabilities of existing models.

\section{Conclusion}

In this paper, we introduced ChatMotion, a sophisticated multi-agent framework that integrates large language models with specialized motion-analysis modules to address the limitations inherent in single-model systems. By dynamically breaking down complex tasks, aggregating diverse model outputs, and carefully selecting the most reliable results, ChatMotion effectively mitigates biases in motion understanding and delivers robust, context-aware analyses. Through experiments conducted on human motion benchmarks such as MoVid-Bench and BABEL-QA, we demonstrated significant improvements in both accuracy and adaptability across various motion tasks.

\clearpage
\bibliography{acl_latex}

\end{document}